\documentclass{article}

\newif\ifbook
\bookfalse

\usepackage[utf8]{inputenc}
\usepackage{stmaryrd}
\usepackage{amsmath,amsthm, amssymb, amsfonts}
\usepackage{fullpage}
\usepackage{footmisc}
\usepackage[figurewithin = section]{caption}
\usepackage{paralist}
\usepackage{wasysym}
\usepackage{verbatim}
\usepackage{authblk}
\usepackage{ltablex}
\usepackage{xcolor}
\usepackage{shadethm}
\usepackage{textcomp}
\usepackage[pdfencoding=auto]{hyperref}
\usepackage{graphicx}
\usepackage{enumitem}
\usepackage{dsfont}
\usepackage{microtype}
\usepackage{bbm}
\usepackage{csquotes}
\usepackage[export]{adjustbox}
\usepackage{comment}
\usepackage{multirow}
\usepackage{makeidx}
\usepackage[ruled,vlined]{algorithm2e}
\usepackage{tikz}
\usetikzlibrary{calc,shapes,matrix,chains,positioning,decorations.pathreplacing,arrows}
\usepackage{mathtools}
\usepackage{wrapfig}
\usepackage{soul}
\DeclareMathSymbol{\mlq}{\mathord}{operators}{``}
\DeclareMathSymbol{\mrq}{\mathord}{operators}{`'}
\DeclareMathOperator*{\Motimes}{\text{\scalebox{0.5}{$\bigotimes$}}}

\definecolor{darkgreen}{HTML}{1BA673}
\definecolor{darkred}{rgb}{0.55, 0.0, 0.0}

\title{Relational Graph Convolutional Networks: A Closer Look}
\author{
Thiviyan Thanapalasingam\thanks{Informatics Institute, University of Amsterdam,
1098XH Amsterdam,The Netherlands}
\qquad 
Lucas van Berkel$^{*}$
\qquad 
Peter Bloem\thanks{Vrije Universiteit Amsterdam, 1081HV Amsterdam, The Netherlands}
\qquad 
Paul Groth$^{*}$
}
\date{}

\begin{document}

\maketitle

\begin{abstract}
In this paper, we describe a reproduction of the Relational Graph Convolutional Network (RGCN). Using our reproduction, we explain the intuition behind the model. Our reproduction results empirically validate the correctness of our implementations using benchmark Knowledge Graph datasets on node classification and link prediction tasks. Our explanation provides a friendly understanding of the different components of the RGCN for both users and researchers extending the RGCN approach. Furthermore, we introduce two new configurations of the RGCN that are more parameter efficient. The code and datasets are available at \url{https://github.com/thiviyanT/torch-rgcn}.
\end{abstract}

\setcounter{tocdepth}{2}

\section{Introduction}\label{intro}

Knowledge Graphs are graph-structured knowledge bases, representing entities and relations between pairs of entities \cite{nickel2015review}. They have become critical for large-scale information systems for tasks ranging from question answering to search \cite{noy2019industry}. The ability to perform statistical relational learning over such data enables new links, properties, and types to be inferred \cite{nickel2015review}, and performing this at a large scale is fundamental for the advancement of the Semantic Web. Additionally, models that can be applied to Knowledge Graphs can also be applied to Relational Database Management Systems, because there is a one-to-one mapping between the two \cite{bornea2013building}.

Relational Graph Convolution Networks (RGCNs) \cite{schlichtkrull2018modeling} are message passing frameworks for learning valuable latent features of relational graphs. RGCNs \cite{schlichtkrull2018modeling} have become been widely adopted for combining Knowledge Graphs with machine learning applications\footnote{The original paper has received over 1200 citations.} for their uses include Knowledge Graph refinement \cite{paulheim2017knowledge}, soft-query answering \cite{daza2020message}, and logical reasoning \cite{sinha2020evaluating}. The original reference code for the RGCN is built on older platforms that are no longer supported. Furthermore, other reproductions \cite{wang2019dgl,fey2019fast} are incomplete.\footnote{At the time of writing, we are aware of additional implementations in PyTorch Geometric \cite{fey2019fast}, which reproduces only the RGCN layer, and in Deep Graph Library \cite{wang2019dgl}, which provides the node classification and link prediction model. However, they are not focused on scientific reproducibility.} Existing descriptions of the model are are mathematically dense, with little exposition, and assume prior knowledge about geometric deep learning.

We reproduced the model in PyTorch, a widely-used framework for deep learning \cite{paszke2019pytorch}. Using our reproduction, we provide a thorough and friendly description for a wider audience. The contributions of this paper are as follows:

\begin{enumerate}
\item A \emph{reproduction of the experiments} as described in the original paper by Schlichtkrull \emph{ et al.} \cite{schlichtkrull2018modeling};
\item A new publicly available PyTorch implementation of the model called \emph{Torch-RGCN};
\item A description of the \emph{subtleties of the RGCN} that are crucial for its understanding, use and efficient implementation; 
\item \emph{New configurations} of the model that are parameter efficient.
\end{enumerate}

The rest of the paper is organized as follows. We begin with an overview of the related work in Section \ref{relatedwork}. Then, we introduce the RGCN in Section \ref{intro-rgcn}, followed by a description of our reimplementation (Section \ref{impl}). We then proceed to describe the reproduction of node classification (Section \ref{entity-class}) and link prediction models with associated experiments (Section \ref{link-pred}). These sections include the model variants mentioned above. In Section \ref{discussion}, we discuss the lessons learned from reproducing the original paper and the implications of our findings. Finally, we discuss our results and conclude.

\section{Related Work}\label{relatedwork}

Machine learning over Knowledge Graphs involves learning low-dimensional continuous vector representations, called Knowledge Graph Embeddings. Here, we briefly review relevant works in the literature.

Commonly, Knowledge Graph Embedding (KGE) models are developed and tested on link prediction tasks. Two key examples of KGEs are TransE \cite{bordes2013translating}, in which relations are translational operations used learned to entities as low-dimensional vectors, and DistMult \cite{yang2014embedding} where the likelihood of a triple is quantified by a multiplicative scoring function. Ruffinelli \emph{et al.} \cite{ruffinelli2019you} provide an overview of the most popular KGE methods and their relative performance. The RGCN is a type of Knowledge Graphs Embedding model. However, RGCNs are different from traditional link predictors, such as TransE and DistMult, because RGCNs explicitly uses nodes' neighborhood information for learning vector representations for downstream tasks \cite{battaglia2018relational}.

Besides RGCNs, there are other graph embedding models for relational data. Relational Graph Attention Networks uses self-attention layers to learn attention weights of edges in relational graphs but yields similar, or in some cases poor, performance when compared to the RGCN. \cite{busbridge2019relational}. Heterogenous Information Network (HIN) exploit meta-paths (a sequence consisting of node types and edge types for modelling particular relationships) to low-dimensional representation of networks \cite{huang2017heterogeneous}. HIN do not use message passing and their expressivity depends on the selected meta-paths.

Beyond node classification and link prediction, RGCN's have other practical uses. For example, Daza \emph{et al.} \cite{daza2020message} explored RGCNs for soft query answering by embedding queries structured as relational graphs which typically consists of a few nodes. Recently, Hsu \emph{et al.} shows that RGCN can be used to extract dependency trees from natural language \cite{guo2021mrgcn}.

\section{Relational Graph Convolutional Network}\label{intro-rgcn}

In \cite{schlichtkrull2018modeling}, Schlichtkrull \emph{et al.} introduced the RGCN as a convolution operation that performs message passing on multi-relational graphs. In this section, we are going to explain the Graph Convolutional \cite{kipf2016semi} and how they can be extended for relational graphs. We will describe message passing in terms of matrix multiplications and explain the intuition behind these operations.

\subsection{Message Passing}

\begin{figure}[ht]
\begin{center}
\includegraphics[width=11cm]{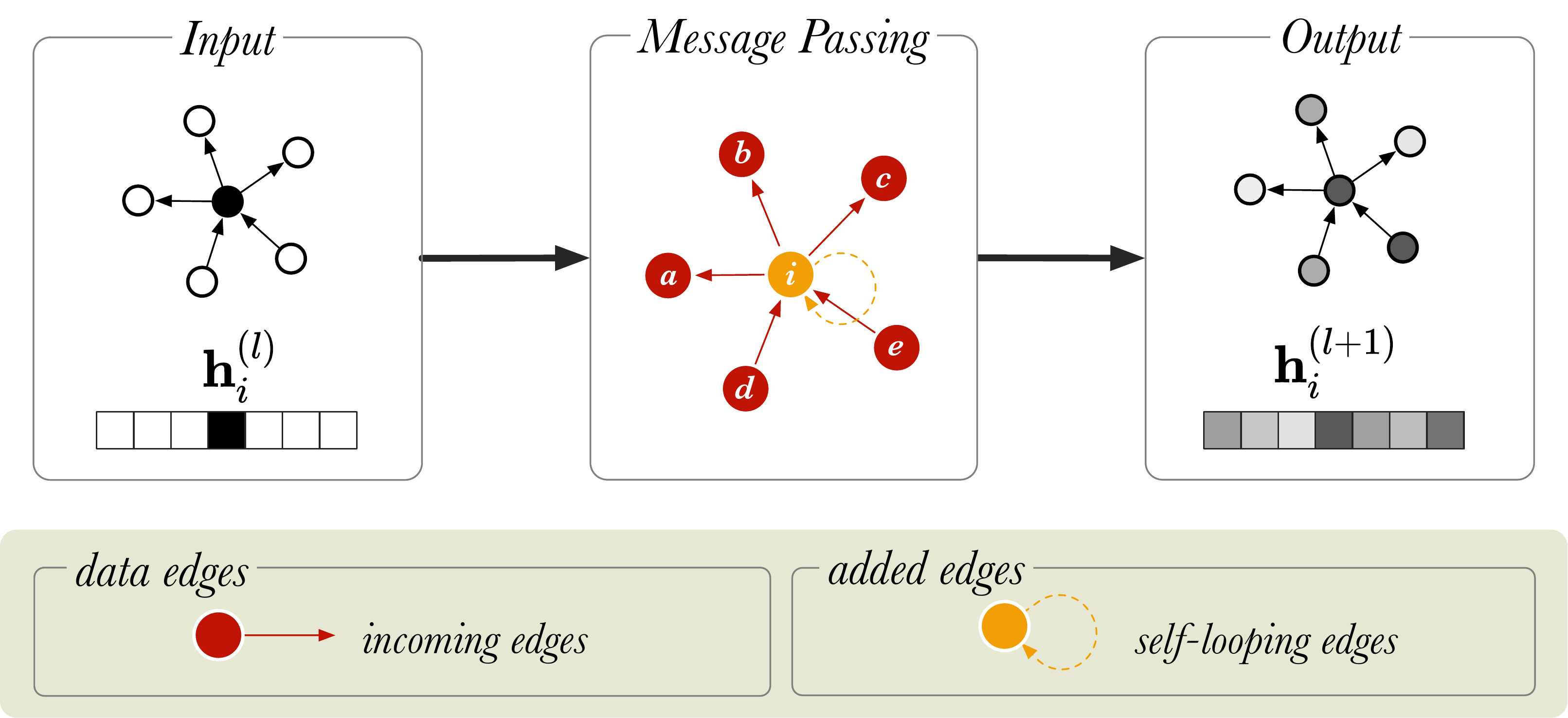}
\caption{A schematic diagram of message passing in a directed graph with 6 nodes. $\textbf{h}_{i}$ is a vector that represents the node embedding of the node $i$ (in orange). $\textbf{h}_{i}^{(l)}$ and $\textbf{h}_{i}^{(l+1)}$ show the node embedding before and after the message passing step, respectively. The neighboring nodes are labelled from $a$ to $e$.}
\label{message passing - gcn}
\end{center}
\end{figure}

We will begin by describing the basic Graph Convolutional Network (GCN) layer for \emph{directed graphs}.\footnote{The original Graph Convolutional Network \cite{kipf2016semi} operates over undirected graphs.} This will serve as the basis for the RGCN in Section \ref{rgcn-section}.

The GCN \cite{kipf2016semi} is a graph-to-graph layer that takes a set of vectors representing nodes as input, together with the structure of the graph and generates a new collection of representations for nodes in the graph. A directed graph is defined as $\mathcal{G}=(\mathcal{V}, \mathcal{E})$, where $\mathcal{V}$ is a set of vertices (nodes) and $\langle i, j\rangle \in \mathcal{E}$ is a set of tuples indicating the presence of \emph{directed} edges, pointing from node $i$ to $j$. Equation \ref{gcn-mp} shows the message passing rule of a \emph{single layer} GCN for an undirected graph, $\mathcal{G}$.

\begin{equation}
\label{gcn-mp}
H =\sigma\left(  A X W \right), 
\end{equation}

Here, $X$ is a node feature matrix, $W$ represents the weight parameters, and $\sigma$ is a non-linear activation function. $A$ is a matrix computed by row-normalizing\footnote{For undirected graphs, a symmetrically normalized Laplacian matrix is used instead \cite{kipf2016semi}.} the adjacency matrix of the graph $\mathcal{G}$. The row normalization ensures that the scale of the node feature vectors do not change significantly during message passing. The node feature matrix, $X$, indicate the presence or absence of a particular feature on a node. 

Typically, more than a single convolutional layer is required to capture the complexity of large graphs. In these cases, the RGCN layers are stacked one after another so that the output of the preceeding RGCN layer $H^{(l-1)}$ is used as the input for the current layer $H^{(l)}$, as shown in Equation \ref{gcn-mp2}. In our work, we will use superscript $l$ to denote the current layer.

\begin{equation}
\label{gcn-mp2}
H^{(l)} =\sigma\left(  A \; H^{(l-1)} \; W \right), 
\end{equation}

If the data comes with a feature vector for each node, these can be used as the input $X$ for the first layer of the model. If feature vectors are not available, one-hot vectors, of length $N$ with the non-zero element indicating the node index, are often used. In this case, the input $X$ becomes the identity matrix $I$, which can then be removed from Equation \ref{gcn-mp}.

We can rewrite Equation \ref{gcn-mp} to make it explicit how the node representations are updated based on a node's neighbors:

\begin{equation}
\label{gcn-mp3}
\textbf{h}_{i} = \sigma\left[\sum_{x \in N_{i}} \frac{1}{\vert N_{i} \vert} {\textbf{x}_i}^{T} W\right].
\end{equation}

Here, $\textbf{x}_i$ is an input vector representing node $i$, $\textbf{h}_{i}$ is the output vector for node $i$, and $N_{i}$ is the collection of the incoming neighbors of $i$, that is the nodes $j$ for which there is an edge ($j$, $i$) in the graph. For simplicity, the bias term is left out of the notation but it is usually included. We see that the GCN takes the average of $i$'s neighbouring nodes, and then applies a weight matrix $W$ and an activation $\sigma$ to the result. Multipliying ${\textbf{x}_i}^{T} W$ by $\frac{1}{\vert N_{i} \vert}$ means that we sum up all the feature vectors of all neighboring nodes. This makes every convolution layer \emph{permutation equivariant}, and that is: if either the nodes (in $A$) are permuted, then the output representations are permuted in the same way. Overall, this operation has the effect of passing information about neighboring nodes to the node of interest, $i$ and this called \emph{message passing}. Message passing is graphically represented in Figure \ref{message passing - gcn} for an undirected graph, where messages from neighboring nodes ($a-e$) are combined to generate a representation for node $i$. After message passing, the new representation of node $i$ is a mixture of the vector embeddings of neighboring nodes. 

If a graph is sparsely connected, a single graph-convolution layer may suffice for a given downstream task. Using more convolutional layers encourages mixing with nodes more than 1-hop away, however it can also lead to output features being oversmoothed \cite{li2018deeper}. This is an issue as the embeddings for different nodes may be indistinguishable from each other, which is not desirable.

In summary, GCNs perform the following two operations: 1) They replace each node representation by the average of its neighbors, and 2) They apply a linear layer with a nonlinear activation function $\sigma$. There are two issues with this definition of the GCN. First, the input representation of node $i$ does not affect the output representation, unless the graph contains a self-loop for $i$. This is often solved by adding self-loops explicitly to all nodes. Second, only the representations of nodes that have incoming links to i are used in the new representation of i. In the relational setting, we can solve both problems elegantly by adding relations to the graph which we will describe in the next section.

\subsection{Extending GCNs for multiple relations}\label{rgcn-section}

In this section, we explain how the basic message passing framework can extended to Knowledge Graphs. We define a Knowledge Graph as a directed graph with labelled vertices and edges. Formally, a KG can be defined as $\mathcal{G}=(\mathcal{V}, \mathcal{E}, \mathcal{R})$, where $\mathcal{R}$ represents the set of edge labels (relations) and $\langle s, r, o \rangle \in \mathcal{E}$ is a set of tuples representing that a subject node $s$ and an object node $o$ are connected by the labelled edge $r \in \mathcal{R}$.

The Relational Graph Convolutional Network extends graph convolutions to Knowledge Graphs by accounting for the directions of the edges and handling message passing for different relations separately. Equation \ref{rgcn-mp} is an extension of the regular message passing rule (Equation \ref{gcn-mp}).

\begin{equation}
\label{rgcn-mp}
H = \sigma\left(\sum_{r=1}^{R} A_{r} X W_{r}\right),
\end{equation}

\noindent where  $R$ is the number of relations, $A_{r}$ is an adjacency matrix describing the edge connection for a given relation $r$ and $W_{r}$ is a relation-specific weight matrix. The extended message passing rule defines how the information should be mixed together with neighboring nodes in a relational graph. In the message passing step, the embedding is summed over the different relations.\footnote{Since RGCN layers are stacked such that the input of a layer is the ouput of the previous layer, taking the sum over $R$ actually inflats the activations. However, for two-layer networks this does not seem to affect performance. For deeper models, taking the mean over the relations rather than the sum may be more appropriate.} 

With the message passing rule discussed thus far, the problem is that for a given triple $\langle s,r,o\rangle$ a message is passed from $s$ to $o$, but not from $o$ to $s$. For instance, for the triple $\langle \texttt{Amsterdam,$ $located\_in,$ $The\_Netherlands} \rangle$ it would be desirable to update both $\texttt{Amsterdam}$ with information from $\texttt{The\_Netherlands}$, and $\texttt{The\_Netherlands}$ with information from $\texttt{Amsterdam}$, while modelling the two directions as meaning different things. To allow the model to pass messages in two directions, the graph is amended inside the RGCN layer by including inverse edges: for each existing edge $\langle s,r,o\rangle$, a new edge $\langle o,r^{\prime},s\rangle$ is added where $r^{\prime}$ is a new relation representing the inverse of $r$. A second problem with the naive implementation of the (R)GCN is that the output representation for a node $i$ does not retain any of the information from the input representation. To allow such information to be retained, a self-loop $\langle s,r_{s},s\rangle$ is added to each node, where $r_{s}$ is a new relation that expresses identity. Altogether, if the input graph contains $R$ relations, the amended graph contains $2R +1$ relations: $\mathcal{R}^{+} = \mathcal{R} \cup \mathcal{R}^{\prime} \cup \mathcal{R}_{s}$.

\begin{figure}[ht]
\begin{center}
\includegraphics[width=11cm]{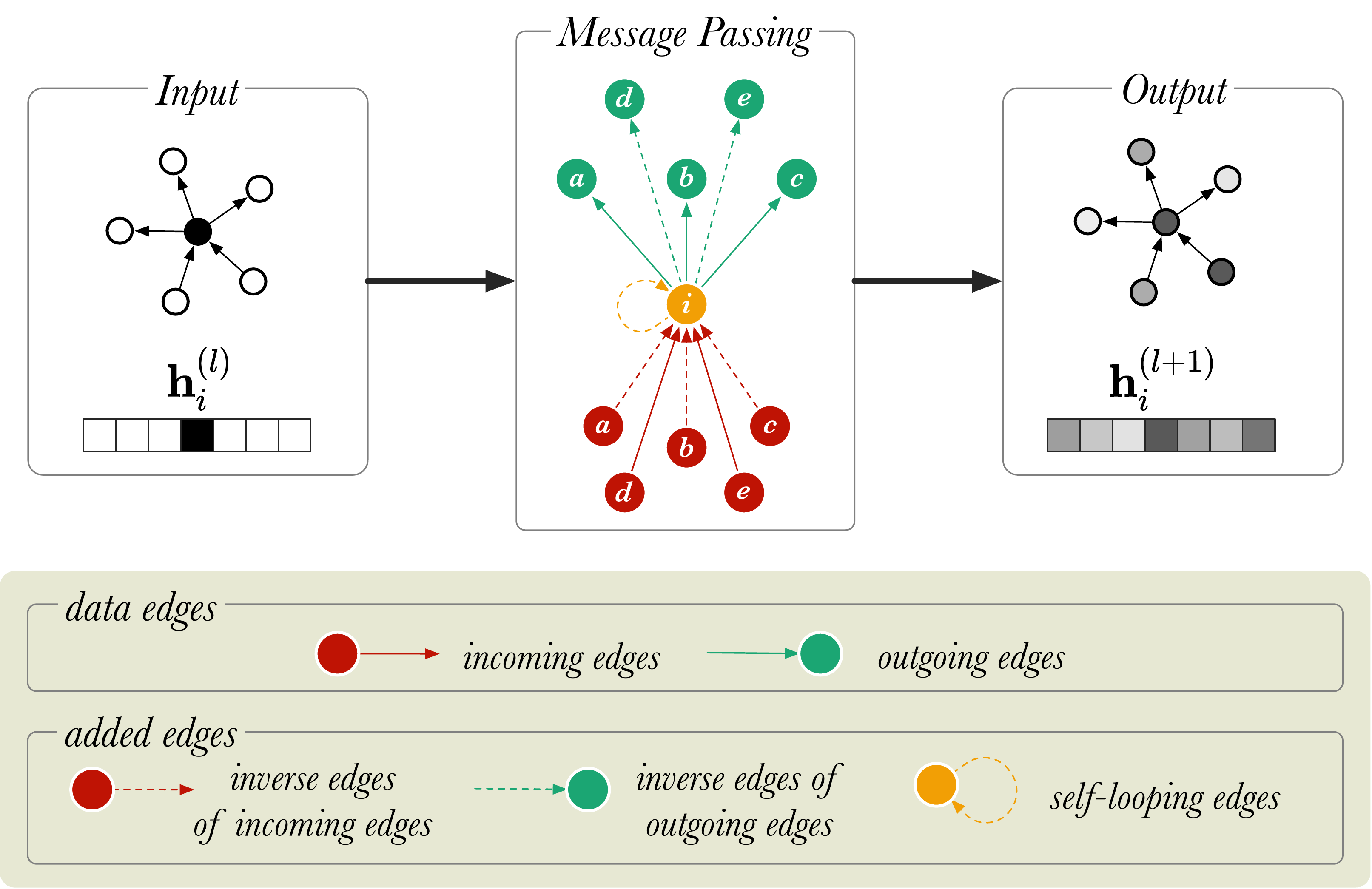}
\caption{A diagram of message passing in a directed, labelled graph with 6 nodes. $\textbf{h}_{i}$ is a vector that represents the node embedding of the node $i$ (in orange). $\textbf{h}_{i}^{(l)}$ and $\textbf{h}_{i}^{(l+1)}$ show the node embedding before and after the message passing step, respectively. The neighboring nodes are labelled from $a$ to $e$.}
\label{rgcn - mp}
\end{center}
\end{figure}

\subsection{Reducing the number of parameters}

We use $N_{in}$ and $N_{out}$ to represent the input and output dimensions of a layer, respectively. While the GCN \cite{kipf2016semi} requires $N_{in} \times N_{out}$ parameters, relational message passing uses $R^{+} \times N_{in} \times N_{out}$ parameters. In addition to the extra parameters required for a separate GCN for every relation, we also face the problem that Knowledge Graphs do not usually come with a feature vector representing each node. As a result, as we saw in the previous section, the first layer of an RGCN model is often fed with a one-hot vector for each node. This means that for the first layer $N_{in}$ is equal to the number of nodes in the graph.

\begin{figure}[ht]
\begin{center}
\includegraphics[width=12cm]{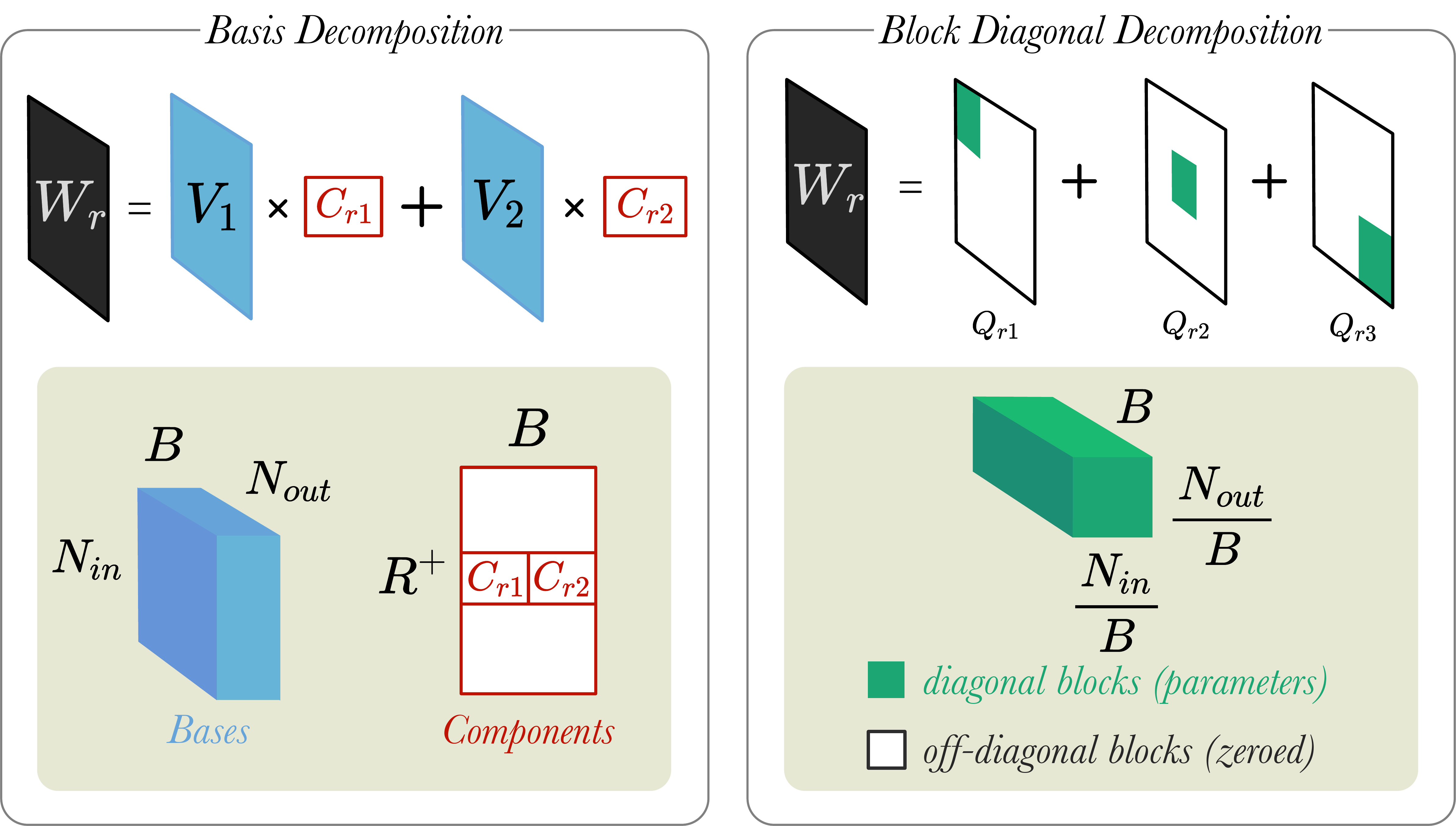}
\caption{A simplified visualisation of the weight regularisation methods. \textbf{Left:} A weight matrix, $W_{r}$, is decomposed into 2 bases ($B=2$). Bases are represented by a tensor $V \in \mathbb{R}^{ B \times N_{in} \times N_{out} }$ and components by matrix $C \in \mathbb{R}^{ R^{+} \times B }$. \textbf{Right:} A different weight matrix, $W_{r}$, is decomposed into 3 blocks ($B=3$). Blocks are tensors $Q_{rb} \in \mathbb{R}^{ {B} \times \frac{N_{in}}{B} \times \frac{N_{out}}{B}}$. $N_{in}$ and $N_{out}$ are the input and output dimensions of layer, respectively.}
\label{fig3}
\end{center}
\end{figure}

In their work, Schlichtkrull \emph{et al.} introduced two different weight regularisation techniques to improve parameter efficiency: 1) Basis Decomposition and 2) Block Diagonal Decomposition. Figure \ref{fig3} shows visually how the two different regularisation techniques work.

\textbf{Basis Decomposition} does not create a separate weight matrix $W_r$ for every relation. Instead, the matrices $W_r$ are derived as linear combinations of a smaller set of $B$ basis matrices $V_b$, which is shared across all relations. Each matrix $W_r$ is then a weighted sum of the basis vectors with component weight $C_{rb}$:

\begin{equation}
  \label{bfd}
  W_{r}=\sum_{b=1}^{B} C_{r b} V_{b},
\end{equation}

Both the component weights and the basis matrices are learnable parameters of the model, and in total they contain fewer parameters than $W_{r}$. With lower number of basis functions, $B$, the model will have reduced degrees of freedom and possibly better generalisation. 

\textbf{Block Diagonal Decomposition} creates a weight matrix for each relation, $W_{r}$, by partitioning $W_{r}$ into $\frac{N_{in}}{B}$ by $\frac{N_{out}}{B}$ blocks, and then fixing the off-diagonal blocks to zeros (shown in Figure \ref{fig3}).\footnote{The off-diagonal blocks are $\frac{N_{in}}{B}$ by $\frac{N_{out}}{B}$ matrices containing only zeros. In Equation \ref{bdd}, we present these zeroed blocks simply with $\emph{0}$.} This deactivates the off-diagonal blocks, such that only the diagonal blocks are updated during training. An important requirement for this decomposition method is that the width/height of $W_r$ need to be divisible by $B$.

\begin{equation}
\label{bdd}
W_{r} = 
\begin{bmatrix}
    Q_{r1} & $\emph{0}$ & \dots &  $\emph{0}$\\
    $\emph{0}$ & Q_{r2} & $\emph{0}$ & $\emph{0}$ \\
    \vdots & $\emph{0}$ & \ddots & \vdots \\
    $\emph{0}$ & $\emph{0}$ & \dots & Q_{rB}
  \end{bmatrix}
\end{equation}

Here, $B$ represents the number of blocks that $W_{r}$ is decomposed into and $Q_{rb}$ are the diagonal blocks containing the relation-specific weight parameters. Equation \ref{bdd} shows that taking the direct sum of $Q_{r}$ over all the blocks gives $W_{r}$, which can also be expressed as the sum of $\text{diag}(Q_{rb})$ over all the diagonal elements in the block matrix. The higher the number of blocks $b$, the lower the number of trainable weight parameters for each relation, $W_{r}$, and vice versa. Block diagonal decomposition is not applied to the weight matrix of the identity relation $r_{s}$, which we introduced in Section \ref{rgcn-section} to add self-loops to the graph.

\section{Torch-RGCN} \label{impl}

The original implementation of the RGCN by Schlichtkrull \emph{ et al.} was written using two different differentiable programming frameworks (Theano 0.9 \cite{al2016theano} and TensorFlow 1.4 \cite{tensorflow2015-whitepaper} \footnote{TensorFlow 2 is not backward compatible with TensorFlow 1 code.}) both which have become obsolete in the past two years. Therefore, we have reproduced the RGCN using PyTorch \cite{paszke2019pytorch}. We will refer to our implementation as \emph{Torch-RGCN} and the original implementation from \cite{schlichtkrull2018modeling} as TensorFlow-RGCN (\emph{TF-RGCN}). Our RGCN implementation is available at \url{https://github.com/thiviyanT/torch-rgcn}.

In this Section, we will describe how we implemented the Relational Graph Convolutional Network. We begin by introducing crucial concepts for the implementation.

\subsection{Einstein Summation}

Message passing requires manipulating high-dimensional matrices and tensors using many different operations (\emph{e.g.} transposing, summing, matrix-matrix multiplication, tensor contraction). We will use Einstein summation to express these operations concisely.

Einstein summation (einsum) is a notational convention for simplifying tensor operations \cite{encyclopediaofmath}. Einsum takes two arguments: 1) an equation\footnote{We use a notation that maps directly to the way einstein summation is used in code, rather than the standard notation.} in which characters identifying tensor dimensions provide instructions on how the tensors should be transformed and reshaped, and 2) a set of tensors that need to be transformed or reshaped. For example,
\text{einsum}($ ik, jk \rightarrow ij, A, B$) represents the following matrix operation:

\begin{equation}
\label{einsum}
C_{ij} = \sum_{k} A_{ik} \cdot B_{jk},
\end{equation}

The general rules of an einsum operation are that indices which are excluded from the result are summed out, and indices which are included in all terms are treated as the batch dimension. We use $\text{einsum}$ operations in our implementation to simplify the message passing operations.

\subsection{Sparsity}

Since many graphs are sparsely connected, their adjacency matrices can be efficiently stored on memory as \emph{sparse tensors}. Sparse tensors are memory efficient because, unlike dense tensors, sparse tensors only store non-zero values and their indices. We make use of sparse matrix multiplications.\footnote{Recent advances in CUDA implementations have made it possible to perform computations involving sparse matrix multiplications to run on the GPU.} For sparse matrix operations on GPUs, the only mulitplication operation that is commonly available is multiplication of a sparse matrix $S$ by a dense matrix $D$, resulting in a dense matrix. We will refer to this operation as spmm($S$, $D$). For our implementation, we endeavour to express the sparse part of the RGCN message passing operation (Equation \ref{rgcn-mp}), including the sum over relations, in a single sparse matrix multiplication.

\subsection{Stacking Trick}

Using nested loops to iteratively pass messages between all neighboring nodes in a large graph would be very inefficient. Instead, we use a trick to efficiently perform message passing for all relations in parallel.

Edge connectivity in a relational graph is represented as a three-dimensional adjacency tensor $ A \in \mathbb{R}^{R^{+} \times N \times N}$, where $N$ represents the number of nodes and $R^{+}$ represents the number of relations. Typically, message passing is performed using batch matrix multiplications as shown in Equation \ref{rgcn-mp}. However, at the time of writing, batch matrix operations for sparse tensors are not available in most Deep Learning libraries. Using spmm is the only efficient operation available, so we  \emph{stack} adjacency matrices and implement the whole RGCN in terms of this operation.

We augment $A$ by stacking the adjacency matrices corresponding to the different relations $A_{r}$ vertically and horizontally into $A_{v} \in \mathbb{R}^{(N + R^{+}) \times N}$ and $A_{h} \in \mathbb{R}^{N \times (N + R^{+})}$, respectively. 

\begin{equation}
\label{}
A_{v}=\left[\begin{array}{l}
A_{1} \\
A_{2} \\
\ \vdots \\
A_{1^{\prime}} \\
A_{2^{\prime}} \\
\ \vdots \\
A_{s}
\end{array}\right]
\end{equation}

\begin{equation}
\label{}
A_{h} = \left[\begin{array}{l}
A_{1}\;A_{2}\;\cdots\;A_{1^{\prime}}\;A_{2^{\prime}}\;\cdots\;A_{s}
\end{array}\right]
\end{equation}

Here, $\left[ \cdot \right]$ represents a concatenation operation, and $A_{v}$ and $A_{h}$ are both sparse matrices. By stacking $A_{r}$ either horizontally or vertically, we can perform message passing using sparse matrix multiplications rather than expensive dense tensor multiplications. Thus, this trick helps to keep the memory usage low. 

Algorithm \ref{algo} shows how message passing is performed using a series of matrix operations. All these are implementations of the same operation, but with different complexities depending on the shape of the input. 

1) If the inputs $X$ to the RGCN layer are one-hot vectors, X can be removed from the multiplication. The \textbf{featureless message passing} simply multiplies  $A$ with $W$, because the node feature matrix $X$ is not given. Note that $X$, in this case, can also be modelled using an identity matrix $I$. However, since $AW=AIW$, we skip this step to reduce computational overhead. 

2) In the \textbf{horizontal stacking approach},  $X$ multiplied with $W$. This yields the $XW$ tensor, which is then reshaped into a $N\mathcal{R} \times N$ matrix. The reshaped $XW$ matrix is then multiplied with $A_{h}$ using spmm.

3) In the \textbf{vertical stacking approach}, the $X$ is mixed with $A_{v}$ using spmm. The product is reshaped into a tensor of dimension $\mathcal{R} \times N \times N$. The tensor $\mathcal{AX}$ is then multiplied with $W$. 

\begin{figure}[h]
\begin{center}
\includegraphics[width=16cm]{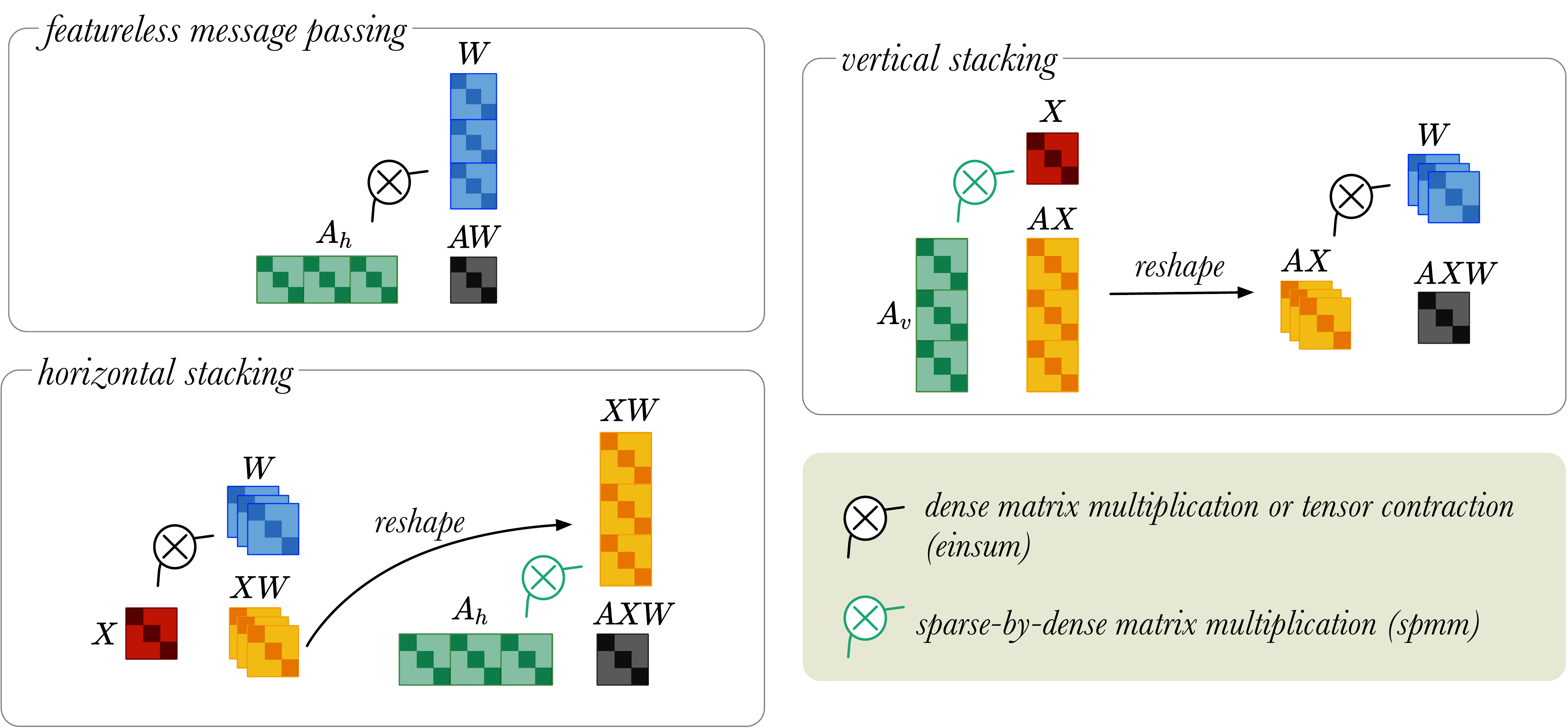}
\caption{ A simplified visual representation of different message passing approaches: (Top-Left) Featureless-message passing, (Bottom-Left) Message passing using horizontally stacked of adjacency matrices, and (Top-Right) Message passing using vertically stacked adjacency matrices. $\bigotimes$ indicates multiplication between dense tensors, which can be implemented with an einsum operator. $\textcolor{darkgreen}{\bigotimes}$ refers to sparse-by-dense multiplication, for which the spmm operation is required. Black arrow indicates tensor reshaping. }
\label{matrix-view}
\end{center}
\end{figure}

Any dense/dense tensor operations can be efficiently implemented with einsum, but sparse/dense operations only allow multiplication of sparse matrix by dense matrix. Therefore, we adopt the two different stacking approaches for memory efficiency. The vertical stacking approach is suitable for low dimensional input and high dimensional output, because the projection to low dimensions is done first. While the horizontal stacking approach is good for high dimensional input and low dimensional output as the projection to high dimension is done last. These matrix operations are visually illustrated in Figure \ref{matrix-view}.

\begin{algorithm}[H]
\label{algo}
\SetAlgoLined
\textbf{Input:} $A$, $\sigma$, [X]\\
\KwResult{$H$}
\eIf{featureless}{
$x \Leftarrow einsum(\mlq ni, io \rightarrow no\mrq, A_h, W)$\\
$H \Leftarrow \sigma\left(x\right)$\\
}{
\eIf{horizontally\_stacked}{
$x \Leftarrow einsum(\mlq ni, rio \rightarrow rno\mrq, X, W)$\\
reshape $x$ into a matrix with the dimensions $NR^{+} \times N_{out}$\\
$x \Leftarrow spmm(A_h, x)$\\
}{
$x \Leftarrow spmm(A_v, X)$\\
reshape $x$ into a tensor with the dimensions $R^{+} \times N \times N_{in}$\\
$x \Leftarrow einsum(\mlq rio, rni \rightarrow no\mrq, W, x)$\\
}
$H \Leftarrow \sigma\left(x\right)$\\
}
 \caption{Message Passing Layer}
\end{algorithm}

Thus far, we focused on how Relational Graph Convolutional layers work and how to implement them. As mentioned in Section \ref{relatedwork}, RGCNs can be used for many downstream tasks. Now, we will discuss how these graph-convolutional layers can be used as building blocks in larger neural networks to solve two downstream tasks implemented in the original RGCN paper \cite{schlichtkrull2018modeling}: \emph{node classification} and \emph{link prediction}. In the next two sections, we detail the model setup, our reproduction experiments and new configurations of the models. We begin with node classification. 

\section{Downstream Task: Node Classification}\label{entity-class}

In the node classification task, the model is trained under a \emph{transductive setting} which means that the whole graph, including the nodes in the test set, must be available during training, with only the labels in the test set withheld. The majority of the nodes in the graph are unlabelled and the rest of the nodes are labelled (we call these \emph{target nodes}). The goal is to infer the missing class information, for example, that \texttt{Amsterdam} belongs to the class \texttt{City}.

\subsection{Model Setup}

\begin{figure}[t]
\begin{center}
\includegraphics[width=16cm]{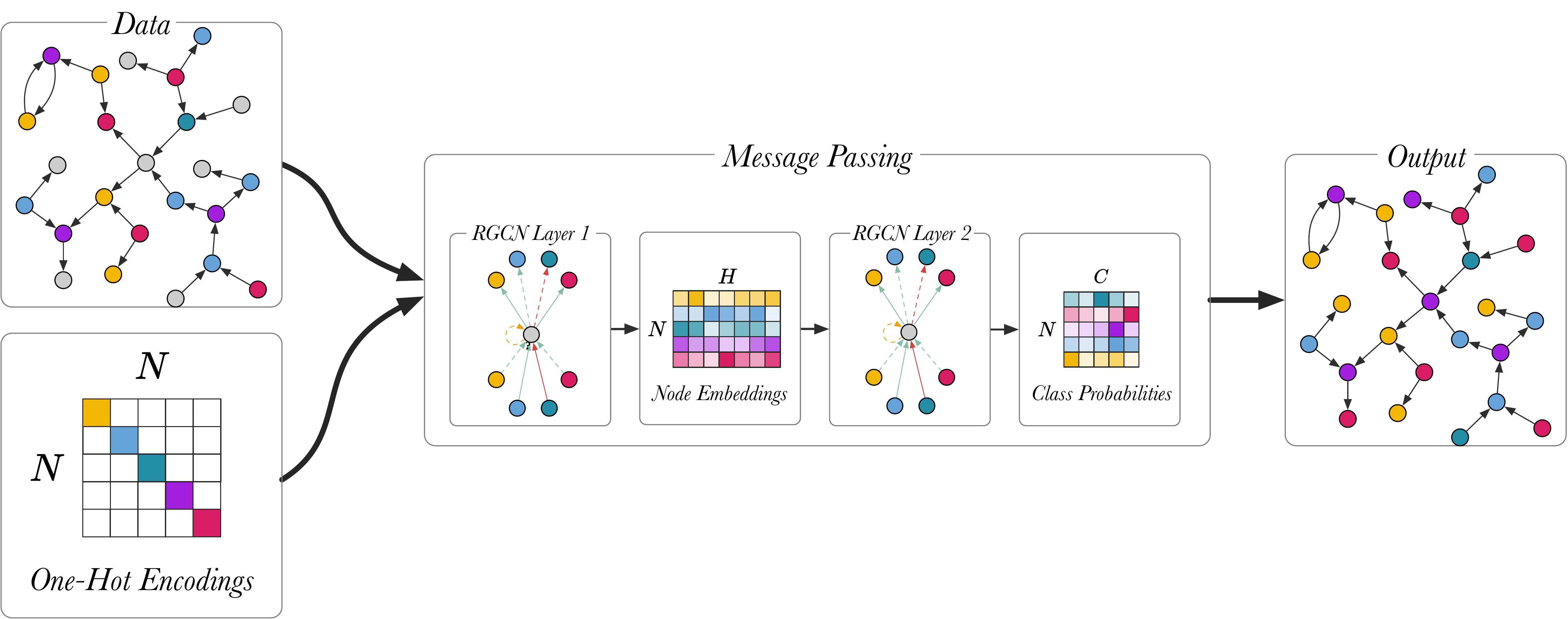}
\caption{ An overview of the node classification model with a two-layer RGCN. Different colour (magenta, green, blue yellow, violet) is used to highlight different entity types. Unlabelled entities are in grey. }
\label{node-classifiction-model}
\end{center}
\end{figure}

Figure \ref{node-classifiction-model} is a diagram of the node classification model with a two-layer RGCN as described in \cite{schlichtkrull2018modeling}. Full-batch training is used for training the node classification model, meaning that the whole graph is represented in the input adjacency matrix $A$ for the RGCN. The input is the unlabeled graph, the output are the class predictions and the true predictions are used to train the model. The first layer of the RGCN is ReLU activated and it embeds the relational graph to produce low-dimensional node embeddings. The second RGCN layer further mixes the node embeddings. Using softmax activation the second layer generates a matrix with the class probabilities, $ Y \in \mathbb{R}^{N \times C}$, and the most probable classes are selected for each unlabelled node in the graph. The model is trained by optimizing the categorical cross entropy loss:

\begin{equation}
\mathcal{L}=-\sum_{i = 1}^{\mathcal{Y}} \sum_{k=1}^{K} t_{i k} \ln \textbf{h}_{i k}^{(L)}, 
\end{equation}

\noindent where $\mathcal{Y}$ is the set of labelled nodes, $K$ represents the number of classes, $t_{i k}$ is one-hot encoded ground truth labels and $\textbf{h}_{i k}^{(L)}$ represents node representations from the RGCN. The last layer ($L$) of the RGCN is softmax-activated. The trained model can be used to infer the classes of unlabelled nodes.

\subsection{e-RGCN}

In GCNs \cite{kipf2016semi}, the node features $X$ are represented by a matrix $X \in \mathbb{R}^{N \times F}$, where $N$ is the number of nodes and $F$ is the number of node features. When node features are not available, one-hot vectors can be used instead. An alternative approach would be to represent the features with continuous values $E \in \mathbb{R}^{N \times D}$, where $D$ is the node embedding dimension.

In the GCN setting \cite{kipf2016semi}, using one hot vectors is functionally very similar to using embedding vectors: the multiplication of the one hot vector by the first weight matrix $W$, essentially selects a row of $W$, which then functions as an embedding of that node. In the RGCN setting, the same holds, but we have a separate weight matrix for each relation, so using one-hot vectors is similar to defining a separate node embedding for each relation. When we feed the RGCN a single node embedding for each node instead, we should increase the embedding dimension $D$ to compensate.

Initial experiments showed that this approach slightly underperforms the one-hot approach on the benchmark data used in \cite{schlichtkrull2018modeling}. After some experimentation, we ended up with the following model, which we call the \emph{embedding-RGCN} (e-RGCN). Its message passing rule is described in Equation \ref{ergcn-mp}. The weight matrix is restricted to a diagonal matrix (with all off diagonal elements fixed to zero).\footnote{This is a special case of the block decomposition with 1x1 blocks.} and then the product is multiplied by the adjacency matrix. 

\begin{equation}
\label{ergcn-mp}
h = \sigma\left(\sum_{r=1}^{R} A_{r} \; E \; \text{diag}(\textbf{w}_{r})\right),
\end{equation}

Here, $E$ is the node embeddings broadcasted across all the relations $\mathcal{R}$, $w_{r}$ is a vector containing weight parameters for relation $r$. Here, $\text{diag}(\cdot)$ is a function that takes a vector $L \in \mathbb{R}^{Q}$ as an input and outputs a diagonal matrix $N \in \mathbb{R}^{Q \times Q}$, where the diagonal elements are elements from the original vector $L$.

Using a diagonal weight matrix improves parameter efficiency, while enabling distinction between relations. We created a new node classfication model, where the first layer is an e-RGCN layer and the second layer is a standard RGCN (without regularisation) that predicts class probabilities. This model provides competitive performance with the RGCN, using only 8\% of the parameters.

\subsection{Node Classification Experiments}

All node classification models were trained following Schlichtkrull \emph{et al.} using full-batch gradient descent for 50 epochs. However, we used 100 epochs for e-RGCN on the AM dataset. Glorot uniform initialisation \cite{glorot2010understanding} was used to initialise parameters with a gain of $\sqrt{2}$ corresponding to the ReLU activation function. Kaiming initialization \cite{he2015delving} was used to initialise the node embeddings in the e-RGCN node classification model.\footnote{The gain parameter is taken from the DGL implementation \cite{wang2019dgl}. The original implementation does not appear to apply a gain. This choice does not seem to affect the classification performance.} Basis decomposition was used for the RGCN-based node classfication.\footnote{40 bases for the AM \& BGS. 30 bases for MUTAG.} All RGCN and e-RGCN models, except for the e-RGCN on the AM dataset, were trained using a GPU.
 
\subsubsection{Datasets}

We reproduce the node classification experiments using the benchmark datasets that were used in the original paper: AIFB \cite{bloehdorn2007kernel}, MUTAG \cite{debnath1991structure}, BGS \cite{de2013fast} and AM \cite{de2012supporting}. We also evaluate e-RGCN on the same datasets. AIFB is a dataset that describes a research institute in terms of its staff, research group, and publications. AM (Amsterdam Museum) is a dataset containing information about artifacts in the museum. MUTAG is derived as an example dataset for the machine learning model toolkit about complex molecules. The BGS (British Geological Survey) dataset contains information about geological measurements in Great Britain. 

Since the messages in a two-layer RGCN cannot propagate further than two hops, we can prune away the unused nodes from the graph. This significantly reduces the memory usuage for large datasets (BGS \& AM) without any performance deterioration. To the best of our knowledge, this was first implemented in the DGL library \cite{wang2019dgl}. For the AM and BGS datasets, the graph was pruned by removing any nodes that are 2 hops away from the target nodes. Pruning significantly reduces the number of entities, relations and edges and thus, lowers the memory consumption of the node classification model, making it feasible to train it on a GPU with 12GB of memory.  Table \ref{node-classification-dataset} shows the statistics for  the node classification datasets. We use the same training, validation and test split as in \cite{schlichtkrull2018modeling}.

\begin{table}[ht]
\begin{center}
\caption{ Number of entities, relations, edges and classes along with the number of labeled entities for each of the datasets. \emph{Labeled} denotes the subset of entities that have labels and \emph{entities} are the nodes without any labels. (*) indicates that entities more than two hops away from the target label were pruned.}
\label{node-classification-dataset}

\bigskip
\begin{tabular}{lrrrr}
\hline Dataset & AIFB & MUTAG & BGS* & AM* \\
\hline Entities & 8,285 & 23,644 & 87,688 & 246,728 \\
Relations & 45 & 23 & 70 & 122 \\
Edges & 29,043 & 74,227 & 230,698 & 875,946 \\
Labeled & 176 & 340 & 146 & 1,000 \\
Classes & 4 & 2 & 2 & 11 \\
\hline
\end{tabular}
\end{center}
\end{table}

\subsubsection{Results}

Table \ref{node-classification-results} shows the results of the node classification experiments in comparison to the original RGCN paper. Torch-RGCN achieves similar performances to TF-RGCN reported in \cite{schlichtkrull2018modeling}. We observed that the training times of the node classification models largely depended on the size of the graph dataset. The CPU training times varied from 45 seconds for the AIFB dataset to 20 minutes for the AM dataset. Since our implementation makes use of GPU's, we were able to run the Torch-RGCN models on a GPU and train the model within a few minutes. 

\begin{table}[h]
\begin{center}
\caption{Node classification accuracy for TF-RGCN, Torch-RGCN and e-RGCN. Results for TF-RGCN were taken from the original paper. These are averages over 10 runs, with standard deviations.}
\label{node-classification-results}

\bigskip

\begin{tabular}{rcccc}
\hline
\multirow{2}{*}{Dataset} &  \multicolumn{3}{c}{Model Accuracy (\%)} \\ 
 & TF-RGCN & Torch-RGCN & e-RGCN \\ \hline
AIFB &  95.83 \small{$\pm$0.62} & 95.56 \small{$\pm$0.61} & 89.17 \small{$\pm$0.28} \\
AM & 89.29 \small{$\pm$0.35} & 89.19 \small{$\pm$0.35} & 89.04 \small{$\pm$0.25} \\
BGS & 83.10 \small{$\pm$0.80} & 82.76 \small{$\pm$0.89} & 81.72 \small{$\pm$0.53}  \\
MUTAG & 73.23 \small{$\pm$0.48} & 73.38 \small{$\pm$0.60} & 71.03 \small{$\pm$0.58} \\ \hline
\end{tabular}
\end{center}
\end{table}

\section{Downstream Task: Link Prediction}\label{link-pred}

We now turn towards the second task performed in the original paper, multi-relational link prediction. The aim is to learn a scoring function that assigns true triples high scores and false triples low scores \cite{bordes2013translating}, with the correct triple ranking the highest. After training, the model can be used to predict which missing triples might be true, or which triples in the graph are likely to be incorrect.

\subsection{Model Setup}

\begin{figure}[t]
\begin{center}
\includegraphics[width=16cm]{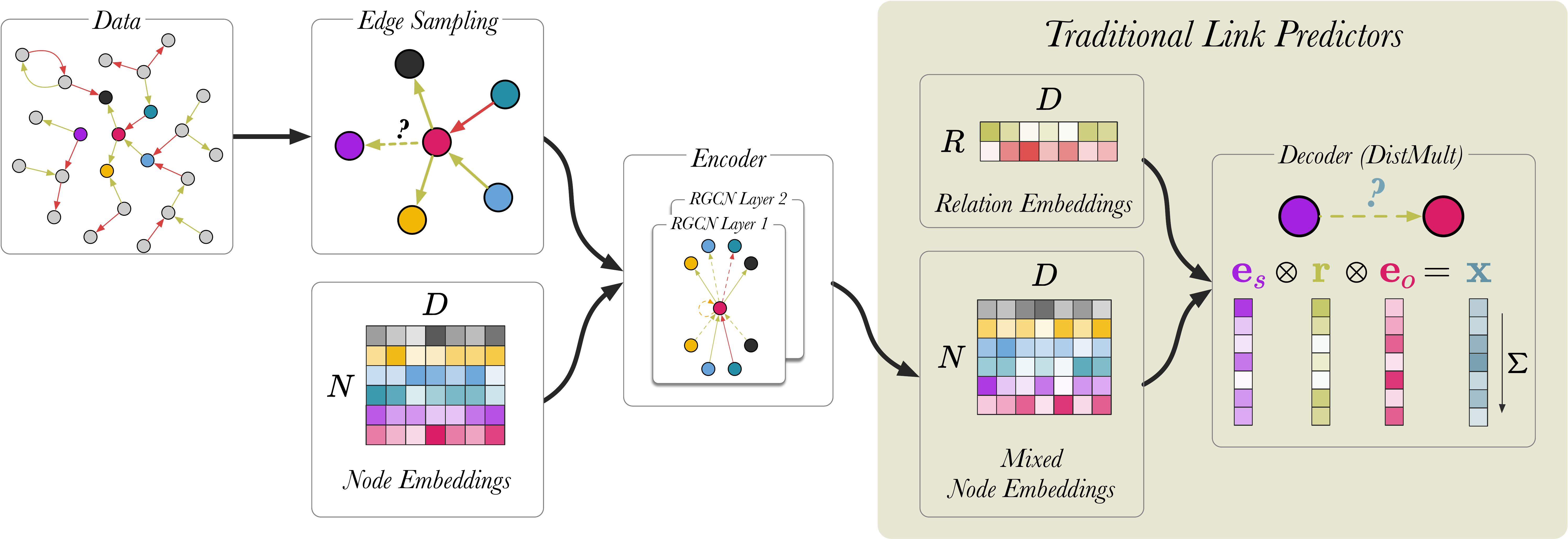}
\caption{ A schematic visualisation of link prediction models. Edges are coloured (red and green) to indicate different edge labels. RGCN-based encoders can be seen an extension to traditional link predictors, such as DistMult \cite{yang2014embedding} and TransE \cite{bordes2013translating}. RGCNs enrich the node representations used by these models by mixing them along the edges of the graph, before applying the score function. In this case, removing the RGCN layers and the upstream edge sampling, recovers the original DistMult. In the last step, the vectors corresponding to entities and relation are element-wise multiplied and the product $x$ is summed. For a given triple $\langle s,r,o \rangle$, the model produces a single scalar value $x$ which indicates how likely the triple is to be true.}
\label{link-prediction-model}
\end{center}
\end{figure}

We follow the procedure outlined by Schlichtkrull \emph{et al.} Figure \ref{link-prediction-model} shows a schematic representation of the link prediction model as described in the original paper. During training, traditional link predictors \cite{bordes2013translating, yang2014embedding} simultaneously update node representations and learn a scoring function (decoder) that predicts the likelihood of the correct triple being true. RGCN-based link predictors introduce additional steps upstream.

We begin by sampling 30,000 edges from the graph using an approach called neighborhood edge sampling (see Section \ref{edge-sampling}). Then, for each triple we generate 10 negative training examples, generating a batch size of 330,000 edges in total. Node embeddings $E \in \mathbb{R}^{N \times D}$ are used an input for the RGCN.\footnote{In the original implementation, the embeddings are implemented as affine operation (\emph{i.e.} biases are included) and they are ReLU activated. We reproduce this behaviour but it is not clear whether this gives any benefits over simple, unactivated embeddings (as used in the e-rgcn).} The RGCN performs message passing over the sampled edges and generates mixed node embeddings. Finally, the DistMult scoring function \cite{yang2014embedding} uses the mixed node embeddings to compute the likelihood of a link existing between a pair of nodes. For a given triple $\langle s,r,o\rangle$, the model is trained by scoring all potential combination of the triple using the function: 

\begin{equation}
\label{distmult}
f(s, r, o) = \sum_{i} \left(\textbf{e}_{s} \Motimes  \textbf{r} \Motimes  \textbf{e}_{o}\right)_{i} = x
\end{equation}

Here, $\textbf{e}_{s}$ and $\textbf{e}_{o}$ are the corresponding node embedding of entities $s$ and $o$, generated by the RGCN encoder. $\textbf{r}$ is a low-dimensional vector of relation $r$, which is part of the DistMult decoder. As Schlichkrull \emph{et al.} highlighted in their work, the DistMult decoder can be replaced by any Knowledge Graph scoring function. We refer the reader to \cite{ruffinelli2019you} and \cite{rossi2021knowledge} for a comprehensive survey of state-of-the-art KGE models.

Similar to previous work on link prediction \cite{yang2014embedding}, the model is trained using negative training examples. For each observed (positive) triple in the training set, we randomly sample 10 negative triples (\emph{i.e.} we use a negative sampling rate of 10). These samples are produced by randomly corrupting either the subject or the object of the positive example (the probability of corrupting the subject is 50\%). Binary cross entropy loss\footnote{Schlichtkrull \emph{et al.} multiply their loss by  $\frac{1}{(1+\omega)|\hat{\mathcal{E}}|}$. $\omega$ is the negative sampling rate and $|\hat{\mathcal{E}}|$ is the number of edges sampled. We leave this term out of our implementation, because it is a constant and thus it would not affect the training.} is used as the optimization objective to push the model to score observable triples higher than the negative ones:

\begin{equation}
\begin{array}{r}
\mathcal{L}=-\sum_{(s, r, o, y) \in \mathcal{T}} y \log l(f(s, r, o))+ (1-y) \log (1-l(f(s, r, o))),
\end{array}
\end{equation}

\noindent where $\mathcal{T}$ is the total set of positive and negative triples, $l$ is the logistic sigmoid function, and $y$ is an indicator set to $y = 1$
for positive triples and $y = 0$ for negative triples. $f(s, r, o)$ includes entity embeddings from the RGCN encoder and relations embeddings from the DistMult decoder. 

\subsubsection{Edge Dropout}

In their work, Schlickrull \emph{et al.} \cite{schlichtkrull2018modeling} apply edge dropout to the link prediction model which acts as an additional regularisation method. This involves randomly selecting edges and removing them from a graph. As described in Section \ref{intro-rgcn}, for every edge in the graph inverse edges $A_{r^{\prime}}$ and self-loops $A_{s}$ are added within the RGCN layer. Dropping edges after this step poses a potential data leakage issue because inverse edges and self-loops of dropped edges will be included in the message passing step and thus, invalidate the model performance. To circumvent this issue, edges are dropped from a graph before feeding it into the RGCN.

Edge dropout is applied such that the dropout rates on the self-loops $\mathcal{R}_{s}$ are lower than for the data edges $\mathcal{R}$ and inverse edges $\mathcal{R}^{\prime}$. One way to think about this that this ensures that the message from a node to itself is prioritised over incoming messages from neighboring nodes. In our implementation, we separate out $A_{s}$ from $A$ and then apply the different edge dropout rates separately. The edge dropout is performed before row-wise normalising $A$.

\subsubsection{Edge Sampling} \label{edge-sampling}

Graph batching is required for training the RGCN-based link prediction model, because it is computationally expensive to perform message passing over the entire graph due to the large number of hidden units used for the RGCN.\footnote{Schlichtkrull \emph{et al.} use a large number of hidden units: 200 for FB15k and WN18; 500 for FB15k-237. In the node classification fewer hidden units are required in the RGCN. This makes the link prediction model more memory demanding.} 

Schlichtkrull \emph{et al.} sample an edge with the probability proportional to its weight. In \emph{uniform edge sampling}, equal weights are given to all the edges. However, in \emph{neighborhood edge sampling}, initial weights are proportional to the node degrees of vertices connected to edges. Then as edges are being sampled, the weight of its neighboring edges is increased and this increases the probability of these edges being sampled \cite{klusowski2018counting}. This sampling approach benefits link prediction because the neighboring edges provide context information to deduce the existence of a relation between a given pair of entities. In contrast, uniform edge sampling assumes that all edges are independent of each other, which is not be applicable to Knowledge Graphs.

\subsection{Link Prediction Experiments}

As in the original paper, the models are evaluated using Mean Reciprocal Rank (MRR) and Hits@$ k $ ($ k $ = 1, 3 or 10). The Torch-RGCN model was trained for 7,000 epochs. We monitored the training of our models by evaluating it at regular intervals every 500 epochs. Schlichtkrull initialisation (\emph{see Appendix}) was used to initialise all parameters in the link prediction models \cite{schlichtkrull2018modeling} and in our reproductions. Schlichtkrull \emph{et al.} \cite{schlichtkrull2018modeling} trained their models on the CPU. Our Torch-RGCN implementation and c-RGCN models run on the GPU. Early stopping was not used. We used the hyperparameters described in \cite{schlichtkrull2018modeling}.

\subsubsection{Datasets}
To evaluate link prediction, Schlichtkrull \emph{et al.} used subsets of Freebase (FB-15k and FB15k-237)\cite{schlichtkrull2018modeling}, and WordNet (WN18) \cite{bordes2013translating}. We only use WN18 \footnote{The link prediction model was expensive to train (3-5 days of training for a single model). WN18 was sufficient to establish reproduction.}. WN18 is a subset of WordNet the graph which describes the lexical relations between words. To check our reproduction, we also used FB-Toy \cite{ruffinelli2019you} which was not in the original paper. FB-Toy is a dataset consisting of a subset of FB15k. In Table \ref{relation-prediction-dataset}, we show the statistics corresponding to these graph datasets.

\subsubsection{Details}

For link prediction, a single-layer RGCN with basis decomposition for WN18 and for FB-Toy a two-layer RGCN with block diagonal decomposition is used. An $\mathcal{L}2$ regularisation penalty of 0.01 for the scoring function is applied. To compute the \emph{filtered} link prediction scores, triples that occur in the training, validation and test are filtered. The Torch-RGCN model is trained on WN18 using batched training, in which 30,000 neighboring edges are sampled at every epoch. An edge dropout rate of 0.2 is applied for self-loops and 0.5 for data edges and inverse edges. Edges are randomly sampled from a Knowledge Graph using the \emph{neighborhood edge} approach. In our reproduction attempts, we have found that this approach enables the model to perform better than uniform edge sampling. The RGCN is initialised using Schlichtkrull normal initialisation (\emph{see Appendix}), while the DistMult scoring function is initialised using standard normal initialisation.

We follow the standard protocol for link prediction. See \cite{ruffinelli2019you} for more details. Some of the hyperparameters used in training were not detailed in \cite{schlichtkrull2018modeling}. To the furthest extent possible, we followed the same training regime as the original paper and code base, and we recovered missing hyperparameters. The hyperparameters for all experiments are provided in documented configuration files on \url{https://github.com/thiviyanT/torch-rgcn}.

\begin{table}[ht]
\begin{center}
\caption{ Number of entities and relation types along with the number of edges per split for the three datasets. }
\label{relation-prediction-dataset}

\bigskip
\begin{tabular}{lrrrr}
\hline Dataset & WN18 & FB-Toy \\
\hline Entities & 40,943 & 280 \\
Relations & 18 & 112 \\
Train edges & 141,442 & 4565 \\
Val. edges & 5,000 & 109 \\
Test edges & 5,000 & 152 \\
\hline
\end{tabular}
\end{center}
\end{table}

\subsubsection{Results}

We verify the \emph{correctness} of our implementation by reproducing the performance on a small dataset (FB-Toy) and by comparing the statistics of various intermediate tensors in the implementation with those of the reference implementation.\footnote{https://github.com/MichSchli/RelationPrediction} We selected a number of intermediate tensors in the link prediction model in our implementation and the original implementation. Then, we measured the statistics of the intermediate tensors. In Table \ref{intermediate-product-statistics} we report the statistics of the intermediate tensors for  TF-RGCN (original model) and Torch-RGCN (our implementation) link prediction models. These results suggests that the parameters used by both models came from a similar distribution and thus verified that they are one-to-one replication.

\begin{table}[ht]
\begin{center}
\caption{ Parameter statistics for intermediate products at various points in link prediction model for the FB-Toy dataset. We report the minimum, maximum, mean and the standard deviation of the distributions. ($\dagger$) Schlichkrull \emph{et al.} used separate weight matrices for data edges $\mathcal{R}$ and inverse edges $\mathcal{R}^{\prime}$, resulting in two intermediate tensors with the dimensions $112 \times 100 \times 5 \times 5$. Thus, we report the statistics for these two tensors.}
\label{intermediate-product-statistics}

\bigskip
\resizebox{\textwidth}{!}{\begin{tabular}{ccccccc}
\hline 
\multirow{2}{*}{No.} & \multirow{2}{*}{Intermediate Tensors} & \multirow{2}{*}{Tensor Dimensions} & &\multicolumn{2}{c}{Parameter Statistics} \\ 
& & & & TF-RGCN & Torch-RGCN \\
\hline 

\multirow{4}{*}{1} & \multirow{4}{*}{Node Embeddings: Initialisation} & \multirow{4}{*}{$280 \times 500$} & min & -0.45209 & -0.48939 \\ 
& & & max & 0.45702 & 0.48641 \\ 
& & & mean & $-0.00004$ & $0.00029$ \\
& & & std & $0.10697$ & $0.10765$ \\
\hline 

\multirow{4}{*}{2} & \multirow{4}{*}{Node Embeddings: Output} & \multirow{4}{*}{$280 \times 500$} & min & $0.0$ & $0.0$ \\ 
& & & max & $0.47124$ & $0.51221$ \\ 
& & & mean & $0.04318$ & $0.04290$ \\
& & & std & $0.06301$ & $0.06273$ \\
\hline 

\multirow{4}{*}{3} & \multirow{4}{*}{RGCN Layer 1: Block Initialisation for data edges and inverse edges $\dagger$} & \multirow{4}{*}{$224 \times 100 \times 5 \times 5$} & min & -1.38892 \& -1.20551 & -1.28504\\ 
& & & max & 1.47686 \& 1.3872786 & 1.26404\\ 
& & & mean & 0.00007 \& -0.00008 & $0.00001$ \\
& & & std & 0.27726 \& 0.27692 & $0.27715$ \\
\hline 

\multirow{4}{*}{4} & \multirow{4}{*}{RGCN Layer 1: Block Initialisation for self-loops} & \multirow{4}{*}{$500 \times 500$} & min & -1.23380 & -1.20324\\ 
& & & max & 1.30949 & 1.16375\\ 
& & & mean & $-0.00049$ & $-0.00095$ \\
& & & std & $0.27716$ & $0.27755$ \\
\hline 

\multirow{4}{*}{5} & \multirow{4}{*}{RGCN Layer 1: Output} & \multirow{4}{*}{$280 \times 500$} & min & $-2.58617$ & $-2.75152$\\ 
& & & max & $2.43774$ & $2.63124$\\ 
& & & mean & $0.02317$ & $0.00799$ \\
& & & std & $0.51760$ & $0.53759$ \\
\hline 

\multirow{4}{*}{6} & \multirow{4}{*}{DistMult: Relation Initialisation} & \multirow{4}{*}{$112 \times 500$} & min & -4.12359 & -3.97444\\ 
& & & max & 4.89700 & 3.95794\\ 
& & & mean & $-0.00947$ & $-0.00186$ \\
& & & std & $0.99675$ & $0.99851$ \\
\hline 

\multirow{4}{*}{7} & \multirow{4}{*}{DistMult: Output} & \multirow{4}{*}{$3300 \times 1$} & min & $-27.21030$ & -30.75097\\ 
& & & max & $27.0885849$ & $25.89389$\\ 
& & & mean & $0.03524$ & $0.78507$ \\
& & & std & $7.75823$ & $7.27595$ \\

\hline
\end{tabular}}
\end{center}
\end{table}

After confirming that our reproduction is correct, we attempted to replicate the link prediction results on the WN18 dataset.\footnote{Runs on FB15k and FB15k-237 took from three to five days to complete training.} Table \ref{relation-prediction-results} also shows the results of the link prediction experiments carried out. The scores obtained by the Torch-RGCN implementation is lower than that of the TF-RGCN model and therefore, we were unable to duplicate the exact results reported in the original paper \cite{schlichtkrull2018modeling}. We believe that the discrepancies between the scores is caused by the differences in the hyperparameter configurations. The exact hyperparameters that Schlichtkrull \emph{et al.} used in their experiments were not available. 

\begin{table}[ht]
  \begin{center}
  \caption{Mean Reciprocal Rank (MRR) and Hits@k (k=1, 3 and 10) for link prediction using RGCN, Torch-RGCN and c-RGCN (see Section \ref{c-rgcn-intro}). Triples from the truth set (train, validation and test set) have been filtered. Results for TF-RGCN was taken from the original paper.}
  \label{relation-prediction-results}
  
  \bigskip
  
  \begin{tabular}{rlcccc}
  \hline
  Dataset & Model &  MRR & Hits@1 & Hits@3 & Hits@10 \\ \hline
  \multirow{3}{*}{WN18} 
  & TF-RGCN & 0.814 & 0.686 & 0.928 & 0.955 \\
  & Torch-RGCN & 0.749 & 0.590 & 0.908 & 0.939 \\
  & c-RGCN & 0.717 & 0.558 & 0.867 & 0.933 \\ \hline
  \end{tabular}
  \end{center}
\end{table}

Despite our best efforts, we were unable to reproduce the exact link prediction results reported in the original paper \cite{schlichtkrull2018modeling}. This is due to the multitude of hyperparameters\footnote{There are at least 10 non-trivial hyperparameters: Number of Epochs, Learning Rate, Graph Batch Size, Negative Sampling Rate, Number of RGCN layers, Dimension of RGCN layers, Weight Decomposition Method, Number of Blocks or number of basis functions, Edge Dropout Rate, $\mathcal{L}2$ regularisation penalty for the scoring function.}, not all of which are specified in the paper, and the long time required to train the model, with runtimes of several days. We did however manage to show the correctness of our implementation using a small-scale experiment. We consider this an acceptable limitation of our reproduction, because the current training time of the RGCN, compared to the state-of-the-art KGE models \cite{ruffinelli2019you}. A Distmult embedding model can be trained in well under an hour on any of the standard benchmarks, and as shown in \cite{ruffinelli2019you}, outperforms the RGCN by a considerable margin. Thus, the precise link prediction architecture described in \cite{schlichtkrull2018modeling} is less relevant in the research landscape.

\subsection{c-RGCN} \label{c-rgcn-intro}

\begin{figure}[t]
\begin{center}
\includegraphics[width=16cm]{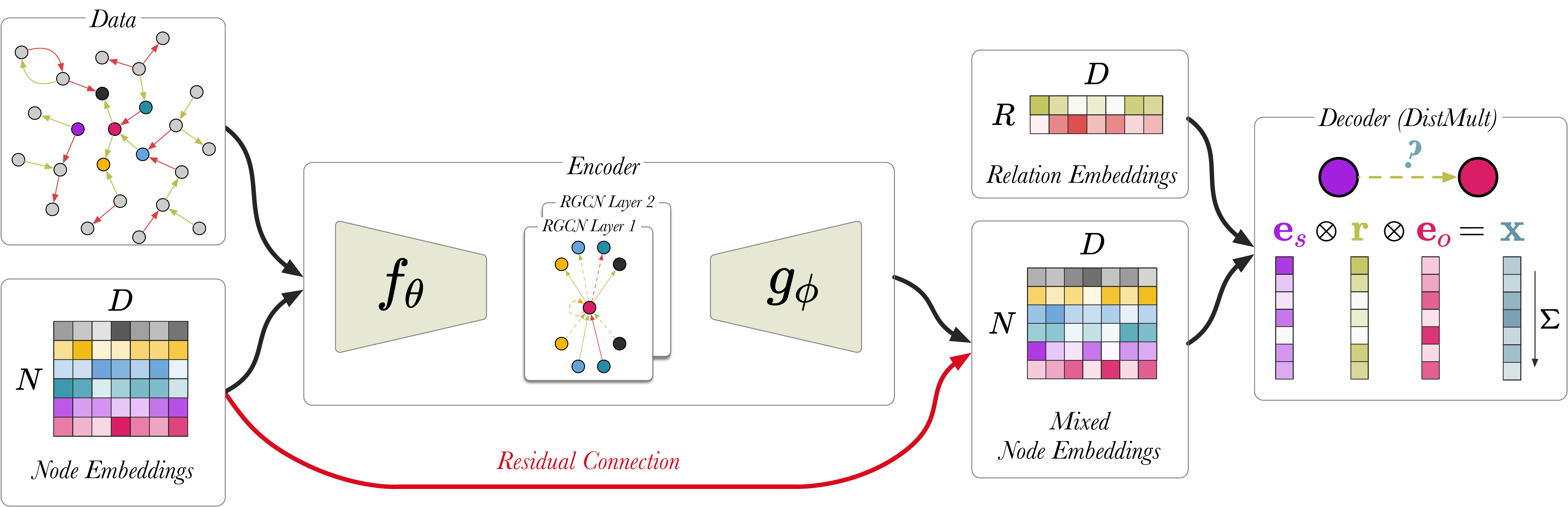}
\caption{ A schematic visualisation of c-RGCN based link prediction model. Here, the encoder has a bottleneck architecture. $f_{\theta}$ and $g_{\phi}$ are linear layers. Prior to message passing $f_{\theta}$ compresses the input node embeddings, and then $g_{\phi}$ projects the mixed node embeddings back up to their original dimensions. The red arrow indicates the residual connection. All edges from the training set are used (\emph{i.e.} edge sampling is not required). }
\label{c-rgcn-link-prediction-model}
\end{center}
\end{figure}

The link prediction architecture presented in \cite{schlichtkrull2018modeling} does not represent a realistic competitor for the state of the art and is very costly to use on large graphs. Furthermore, a problem with the original RGCN link predictor \cite{schlichtkrull2018modeling} is that we need high dimensional node representations to be competitive with traditional link predictors, such as DistMult \cite{yang2014embedding}, but the RGCN is expensive for high dimensions. However, we do believe that the basic idea of message passing is worth exploring further in a link prediction setting. 

To show that there is promise in this direction, we offer a simplified link prediction architecture that uses a fraction of the parameters of the original implementation \cite{schlichtkrull2018modeling} uses. This variant places a bottleneck architecture around the RGCN in the link prediction model, such that the embedding matrix $E$ is projected down to a lower dimension, $C$, and then the RGCN performs message propagation using the compressed node embeddings. Finally, the output is projected up back to the original dimension, $D$, and computes the DistMult score from the resulting high-dimensional node representations. We call this encoding network the \emph{compression-RGCN} (c-RGCN). Equations \ref{encoding-layer} and \ref{decoding-layer} show the message passing rule for the first and second layer of the c-RGCN encoder, respectively. We selected a node embedding size of 128 and compressed it to a vector dimension of 16. We also include a residual connection by including $E$ in the second layer of the c-RGCN. The residual connection allows the model, in principle, to revert back to DistMult if the message passing adds no value. If all RGCN weights are set to 0, we recover the original DistMult. 

\begin{equation}
\label{encoding-layer}
H^{1}=\sigma\left(\sum_{r=1}^{R^{+}} A \; W_{r} \; f_{\theta}( E ) \right), 
\end{equation}

\noindent where $f(X) = XW_{\theta} + b$ with $W_{\theta} \in \mathbb{R}^{C \times E}$.

\begin{equation}
\label{decoding-layer}
H^{2}=E + g_{\phi}\left(\sigma\left(\sum_{r=1}^{R^{+}} A \; W_{r} \; H^{1} \right)\right),
\end{equation}

\noindent where $g(X) = XW_{\phi} + b$ with $W_{\phi} \in \mathbb{R}^{E \times C}$.

As shown in Table \ref{relation-prediction-results}, the c-RGCN does not perform much worse than the original implementation. However, it is much faster, and memory efficient enough for full batch evaluation on the GPU. There is a clear trade-off between compression size of the node embeddings and the performance in link prediction. While this result is far from a state-of-the-art model, it serves as a proof-of-concept that there may be ways to configure RGCN models for a better performance/efficiency tradeoff. 

\section{Discussion} \label{discussion}

We now discuss the implications for the use of the RGCN model, the performance of the new variants and the lessons learned from this reproduction.

\subsection{Implications for RGCN usage}

We believe that Relational Graph Convolutional Networks are still very relevant because it is one of the simplest members of the message passing models and is good starting place for exploration of machine learning for Knowledge Graphs. 

RGCNs clearly perform well on node classification tasks because the task of classifying nodes benefits from message passing. This means that a class for a particular node is selected by reasoning about the classes of neighboring nodes. For example, a researcher can be categorised into a research domain by reasoning about information regarding their research group and close collaborators. Traditional Knowledge Graph Embeddings (KGE) models, such as TransE and DistMult, lack the ability to perform node classification.

While the RGCN is a promising framework, in its current setting we found that the link prediction model proposed by Schlichtkrull \emph{et al.} is not competitive with current state-of-the-art \cite{ruffinelli2019you} and the model is too expensive with considerably lower performance. In our paper, we clarify that RGCN-based link predictors are extensions of KGE models \cite{ruffinelli2019you}, thus training RGCN to predict links will always be more expensive than using a state-of-the-art KGE model. RGCN-based link predictor take several days to train, while state-of-the-art relation models run in well under an hour \cite{ruffinelli2019you}.

To aid the usage of RGCN, we presented two new configurations of the RGCN:

\textbf{e-RGCN.} We propose a new variant of the node classification model which uses significantly less parameters by exploiting a diagonal weight matrix. Our results show that it performs competitively with the model from \cite{schlichtkrull2018modeling}.

\textbf{c-RGCN.} We also present a proof-of-concept model that performs message passing over compressed graph inputs and thus, improves the parameter efficiency for link prediction. The c-RGCN has several advantages over the regular RGCN link predictor: 1) c-RGCN does not require sampling edges, because it is able to process the entire graph in full-batch, 2) c-RGCN takes a fraction of the time it takes to train an RGCN link predictor, 3) c-RGCN uses fewer parameters, and 4) It is straightforward to implement. Although the results for the c-RGCN are not as strong, this sets a path for further development towards efficient message models for relational graphs.

\subsection{Reproduction}

Evolving technologies pose several challenges for the reproducibilty of research artifacts. This includes frequent updates being made to existing frameworks, such as PyTorch and TensorFlow, often breaking backward compatibility. We were in a strong position to execute this reproduction: 1) an author of this paper also worked on the original paper, 2) we contacted one of the lead authors of this paper who was very responsive and 3) we were able to run the original source code \footnote{https://github.com/tkipf/relational-gcn \& https://github.com/MichSchli/RelationPrediction} inside a virtual environment. Nevertheless, we found it considerably challenging to make a complete reproduction. To explain why and to contribute to avoiding such situations in the future, we briefly outline the lessons we have learned during the reproduction. 

\textbf{Parameter Statistics.} There were discrepancies between the description of the link prediction model in \cite{schlichtkrull2018modeling} and the source code. The source code reproduces the values similar to the MRR scores reported in \cite{schlichtkrull2018modeling}. Thus, to reproduce the results we had to perform a deep investigation of the source code. Using the original source, we relied on comparing the parameter statistics and tensor sizes at various points in both models. Since these statistics are helpful to verify the correctness of an implementation, we believe this is a useful practice in aiding reproduction. For complex models with long runtimes, an overview of descriptive statistics of parameter and output tensors for the first forward pass can help to check implementation without running full experiments. We are publishing statistics for intermediate products that we obtained for the link prediction models (see Table \ref{intermediate-product-statistics}).

\textbf{Small dataset.} We found that the link prediction datasets used by \cite{schlichtkrull2018modeling} were large and thus, impractical for debugging RGCN because it is costly to train them on large graphs. Using a smaller dataset (FB-Toy \cite{schlichtkrull2018modeling}) would enable quicker testing with less memory consumption. Thus, we report link prediction results on the FB-Toy dataset \cite{ruffinelli2019you} (Table \ref{c-rgcn-link-prediction-model}).

\textbf{Training times.} The training times were variable and strongly depended on the size of the graph, the number of relations and the number of epochs. Schlichtkrull \emph{et al.} reported the computational complexity, but not practical training times. It turns out that this is an important source of uncertainty in verifying whether re-implementations are correct. We measured the runtimes, which includes training the model and using the pre-trained model for making inference. For 7000 epochs, the link prediction runtimes for Torch-RGCN and c-RGCN on the WN18 dataset are 2407 and 53 minutes, respectively. Node classification experiments took a few minutes to complete, because they only required 50-100 epochs. We encourage authors to report such concrete training times.

\textbf{Hyperparameter Search.} We found that hyperparameters reflect the complexity of the individual datasets. For example, AIFB, the smallest dataset, was not prone to overfitting. Whereas, the larger AM dateset required basis decomposition and needs a reduced hidden layer size. For link prediction, we were unable to identify the optimum hyperparameters for WN18, FB15k and FB15k-237 due to the sheer size of the hyperparameter space and long training times. We provide a detailed list of hyperparameter we use in our reproduction. While this is becoming more common in the literature, this serves as further evidence of the importance of this detailed hyperparameter reporting. 

\textbf{Other factors.} We still faced the common challenges in software reproduction that others have long noted \cite{fokkens-etal-2013-offspring}, including missing dependencies, outdated source code, and changing libraries. An additional challenge with machine learning models is that hardware (e.g. GPUs) now also can impact the performance of the model itself. For instance, while we were able to run the original link prediction code in TensorFlow 1.4, the models no longer seemed to benefit from the available modern GPUs. Authors should be mindful that even if legacy code remains executable for a long time, executing it efficiently on modern hardware may stop being possible much sooner. Here too, reporting results on small-scale experiments can help to test reproductions without the benefit of hardware acceleration.

\section{Conclusion}

We have presented a reproduction of Relational Graph Convolutional Networks and, using the reproduction, we provide a friendly explanation of how message passing works. While message passing is evidently useful for node classification, our findings also show that RGCN-based link predictors are currently too costly to make for a practical alternative to the state-of-the-art. However, we believe that improving the parameter efficiency RGCNs could potentially make it more accessible. We present two novel configurations of the RGCN: 1) e-RGCN, which introduces node embeddings into the RGCN using fewer parameters than the original RGCN implementation, and 2) c-RGCN, a proof-of-concept model which compresses node embeddings and thus speeds up link prediction. These configurations provide the foundation for future work. We believe that the techniques proposed in this paper may also be important for others implementing other message passing models. Lastly, our new implementation of RGCN using PyTorch, TorchRGCN, is made openly available to the community. We hope that this can help serve the community in the use, development and research of this interesting model for machine learning on Knowledge Graphs.

\section*{Acknowledgement}
We are very grateful to Michael Schlichtkrull for supporting us with the reproduction of the link prediction results. Experiments were run on DAS-5 ASCI Supercomputer \cite{bal2016medium} and on the Dutch national e-infrastructure with the support of SURF Cooperative.

\newcommand{\etalchar}[1]{$^{#1}$}
\providecommand{\bysame}{\leavevmode\hbox to3em{\hrulefill}\thinspace}
\providecommand{\MR}{\relax\ifhmode\unskip\space\fi MR }
\providecommand{\MRhref}[2]{%
  \href{http://www.ams.org/mathscinet-getitem?mr=#1}{#2}
}
\providecommand{\href}[2]{#2}

\section*{Appendix}

\subsection*{Notation}

We use lowercase letters to denote scalars (\emph{e.g.} $x$), bold lowercase letters for vectors (\emph{e.g.} $\textbf{w}$), uppercase letters for matrices (\emph{e.g.} $A$), and caligraphic letters for sets (\emph{e.g.} $\mathcal{G}$). We also use uppercase letters for referring to dimensions and lowercase letters for indexing over those dimensions (\emph{e.g.} $\sum_{b=1}^{B}$).

\subsection*{General Experimental Setup}

All experiments were performed on a single-node machine with an Intel(R) Xeon(R) Gold 5118 (2.30GHz, 12 cores) CPU and 64GB of RAM. GPU experiments used a Nvidia GeForce GTX 1080 Ti GPU.  We used the Adam optimiser \cite{kingma2014adam} with a learning rate of 0.01. For reproducibility, we provide an extension description of the hyperparameters that we have used in node classifications and link predictions in YAML files under the \emph{configs} directory on the project GitHub page: \url{https://github.com/thiviyanT/torch-rgcn}. 

We ran the original implementation of the link prediction model \footnote{\label{github-pa}https://github.com/MichSchli/RelationPrediction} on the FB15-237 dataset. The exact hyperparameters for WN18 and FB15k experiments were not available in the original codebase. We used Tensorflow 1.4, Theano 1.0.5 and CUDA 8.0.44. This replication required a Nvidia Titan RTX GPU with 24GB of GPU memory, but model training and inference was performed on the CPU.

\subsection*{Schlichtkrull Initialisation}

The initialization used in the link prediction models in \cite{schlichtkrull2018modeling} differs slightly from the more standard Glorot initialization \cite{glorot2010understanding}.

\begin{equation}
\label{schlichtkrull-init}
std = \Phi \times \frac{3}{\sqrt{\text{fan in} + \text{fan out}}}, 
\end{equation}

Here, gain ($\Phi$) is a constant that is used to scale the standard deviation according the applied non-linearity. $std$ is used to sample random points from either a standard normal or uniform distribution. We refer to this scheme as \emph{Schlichtkrull initialisation}. When gain is not required, $\Phi$ is set to $1.0$.

\subsection*{FB-Toy Link Prediction}

We also performed link prediction experiment on the FB-Toy dataset. The mean and standard error of the link prediction results are reported in Table \ref{fbtoy}.

\begin{table}[ht]
\begin{center}
\caption{Mean and Standard Error of Mean Reciprocal Rank (MRR) and Hits@k (k=1, 3 and 10) over 10 runs for link prediction using RGCN and Torch-RGCN on FB-Toy dataset. Triples from the truth set (train, validation and test set) have been filtered. All models that were trained on the GPU.}
\label{fbtoy}
\bigskip
\begin{tabular}{rlcccc}
\hline
Dataset & Model &  MRR & Hits@1 & Hits@3 & Hits@10 \\ \hline
\multirow{2}{*}{FB15k-Toy} 
& TF-RGCN & $0.432 \pm 0.008$ & $0.293 \pm 0.007$ & $0.482 \pm 0.011$ & $0.768 \pm 0.010$ \\
& Torch-RGCN & $0.486 \pm 0.009$ & $0.352 \pm 0.011$ & $0.540 \pm 0.009$ & $0.799 \pm 0.008$ \\\hline
\end{tabular}
\end{center}
\end{table}

\end{document}